\documentclass{article}

\usepackage[nonatbib, final]{neurips_2021}

\usepackage[utf8]{inputenc} 
\usepackage[T1]{fontenc}    
\usepackage{hyperref}       
\usepackage{url}            
\usepackage{booktabs}       
\usepackage{amsfonts}       
\usepackage{nicefrac}       
\usepackage{microtype}      
\usepackage{xcolor}         
\usepackage{amsmath,amssymb,amsfonts} 

\usepackage[frozencache,cachedir=minted-cache]{minted}
\usepackage{pgfplots}
\usepackage{hyperref}

\usepackage[
backend=biber,
style=numeric,
sorting=none
]{biblatex}

\newcommand{\rmi}{\rm i}

\title{A research framework for writing differentiable PDE discretizations in JAX}

\author{%
  Antonio Stanziola \\
  University College London \\
  \texttt{a.stanziola@ucl.ac.uk} \\
  \And
  Simon R. Arridge \\
  University College London \\
  \texttt{s.arridge@ucl.ac.uk} \\
  \And
  Ben T. Cox \\
  University College London \\
  \texttt{b.cox@ucl.ac.uk} \\
  \And
  Bradley E. Treeby \\
  University College London \\
  \texttt{b.treeby@ucl.ac.uk} \\
}

\begin{document}

\maketitle

\begin{abstract}
Differentiable simulators are an emerging concept with applications in several fields, from reinforcement learning to optimal control. Their distinguishing feature is the ability to calculate analytic gradients with respect to the input parameters. Like neural networks, which are constructed by composing several building blocks called layers, a simulation often requires computing the output of an operator that can itself be decomposed into elementary units chained together. While each layer of a neural network represents a specific discrete operation, the same operator can have multiple representations, depending on the discretization employed and the research question that needs to be addressed. Here, we propose a simple design pattern to construct a library of differentiable operators and discretizations, by representing operators as mappings between families of continuous functions, parametrized by finite vectors. We demonstrate the approach on an acoustic optimization problem, where the Helmholtz equation is discretized using Fourier spectral methods, and differentiability is demonstrated using gradient descent to optimize the speed of sound of an acoustic lens. The proposed framework is open-sourced and available at \url{https://github.com/ucl-bug/jaxdf}
\end{abstract}

\section{Introduction}
In the last decade, the exponential growth of machine learning has been mirrored by a comparable advancement in frameworks for automatic differentiation (AD) and parallelization. This has paved the path for the emergence of a new research field, sometimes named scientific machine learning \cite{rackauckas2020universal}, where elements of machine learning are blended with scientific models and simulators.
In particular, AD, the main algorithm employed for neural network training and generally for analytical gradient estimation, can be used to differentiate with respect to any continuous parameter involved in a simulator \cite{innes2019differentiable, jax2018github}. Conversely, a simulator that allows for AD can be used inside a machine learning model, for example, to implement an implicit layer \cite{chen2018neural} or a physics loss function.

Differentiable simulators are also of broad interest for several fields of research, such as for inverse problems \cite{adler2017solving, adler2017odl}, system identification \cite{jatavallabhula2021gradsim}, optimal control \cite{holl2020learning}, Bayesian inference \cite{cranmer2020frontier}, optimization under uncertainty \cite{gerlach2020koopman}, reinforcement learning \cite{mora2021pods} and also to improve the performance of simulators themselves \cite{heiden2020neuralsim}.  Differentiable simulators can also be used to construct physics losses. There's now a rapidly growing list of differentiable simulators for a large variety of applications, such as molecular dynamics \cite{jaxmd2020}, Finite Element Analysis \cite{yashchuk2020bringing}, computational fluid dynamics \cite{Kochkov2021-ML-CFD}, Physics Informed Neural Networks (PINNs) \cite{lu2021deepxde} and neural differential equations \cite{rackauckas2019diffeqflux}, to name a few.

However, most of the available simulators are often tied to a specific discretization that solves a continuous problem, meaning that different representations require different implementations of the same equations. Furthermore, it is hard to modify the given discretization method for research purposes, at least in ways that the software maintainer does not specifically intend. Additionally, it is of interest for such simulators to be compatible with a machine learning library, to potentially include it as a component of a larger machine learning model.

\paragraph{Aim}
We propose a customizable framework, called jaxdf (Jax- Discretization Framework) for writing differentiable simulators, that decouples the mathematical definition of the problem from the underlying discretization. The underlying computations are performed using JAX \cite{jax2018github}, and are thus compatible with the broad set of program transformations that the package allows, such as AD and batching. This enables rapid prototyping of multiple customized representations for a given problem, to develop physics-based neural network layers and to write custom physics losses, while maintaining the speed and flexibility required for research applications. It also contains a growing open-source library of differentiable discretizations compatible with the JAX ecosystem.

\section{Software model}
Writing a simulator requires a mathematical expression defined over continuous functions, often a Partial Differential Equation (PDE), to be translated into a program that manipulates a finite set of numerical values. This fundamental step is called \textit{discretization}. Here, we assume that functions are defined on a rectangular interval $\Pi \subseteq \mathbb{R}^D$ of the $D$-dimensional Euclidean space.

Given a function $f: \Pi \to \mathbb{R}^M$, a simulation often requires to evaluate and manipulate multiple expressions of the form $\mathbf{L}f$, where $\mathbf{L}$ is a (possibly non-linear) operator acting on the function $f$. To implement this in a computer, a restriction is often made to a parametrized family of functions, such that we can identify elements of this family using a mapping from finite vectors to functions. That is, there's a mapping $\mathcal{D}$ such that
\begin{equation}
    \theta \xrightarrow{\mathcal{D}}f
\end{equation}
with $f \in \text{Range}(\mathcal{D})$ or, in other words, $f_\theta(x) = \mathcal{D}(\theta, x)$ is a function parametrized by $\theta$.
We will call $\mathcal{D}$ the \textit{discretization family} of $f$, while $\theta$ is the \textit{discrete representation} of $f$ over $\mathcal{D}$. The latter is analogous to the interpolation function defined in other libraries.\footnote{See for example the Operator Discretization Library at \url{https://odl.readthedocs.io}}

\paragraph{Examples of discretization families}
A simple example of discretization family is the set of $N$-th order polynomials on the interval $[0,1)$: 
\begin{equation}
    \mathcal{P}_N(\theta,x) = \sum_{i=0}^N \theta_i x^i, \qquad \theta \in \mathbb{R}^{N+1}
\end{equation}
Another common example is the Fourier-spectral representation $\mathcal{\hat F}$ given by the iDFT of the sequence $\theta \in \mathbb{C}^N$, which is defined on a regular collocation grid in frequency space, i.e.
\begin{equation}
\mathcal{\hat F}(\theta, x) = \sum_i^N \theta_i e^{i(k_i \cdot x)}.
\end{equation}
Alternatively, the Fourier representation is given by defining $\theta$ in signal space on a regular spatial collocation grid, using the periodic $\text{sinc}$ function:
\begin{equation}
\mathcal{F}(\theta,x) = \sum_i^N \theta_i \frac{\sin((N+1/2)(x - x_i))}{\sin((x-x_i)/2)}.
\end{equation}
Another important kind of mappings is neural networks, often used in this context to implement Physics-Informed Neural Networks (PINNs). Many other discrete representations are possible, such as Finite Elements, Finite Differences, radial basis functions and splines. In general, the choice of the right representation is application dependent, subject to hardware and software constraints, and often a topic of research on its own.

\subsection{Operator representation}
Let $f_\theta$ be a member of the discretization family $\mathcal{D}$ with parameters $\theta$. Representing the action of an operator $\mathbf{L}$ over the discretization family, means to find the function $\mathcal{G}(\gamma,x)$ with parameters $\gamma$ such that $\mathcal{G}(\gamma,x)$ is close to $\mathbf{L}(f_\theta)$ in some sense. 

In general, the parameters $\theta$ and $\gamma$ can be different. Similarly, also the discretization families $\mathcal{D}$ and $\mathcal{G}$ may be different. 
Rather than implementing the discretized mapping only as an operator acting on the discrete representation, the corresponding discrete operator $\mathbf{\hat L}$ is instead represented as
\begin{equation}
    (\mathcal{D},\theta) \xrightarrow{\mathbf{\hat L}} (\mathcal{G}, \gamma).
\end{equation}
This means that the operator also specifies the output discretization family $\mathcal{G}$. Most often $\mathcal{G}$ is independent of the parameters, and we can split the mapping into two functions
\begin{equation}
    (\mathcal{D},\theta) \xrightarrow{\mathbf{\hat L}} (L_f(\mathcal{D}), L_p(\mathcal{D})(\theta)),
\end{equation}
where $L_f$ is a discrete mapping between discretization families, and $L_p(\mathcal{D})$ is a differentiable function of the parameters $\theta$. Thus, defining an operator $\mathbf{L}$ over a discretization family $\mathcal{D}$, boils down to implementing $L_f(\mathcal{D})$ and $L_p(\mathcal{D})$.

\paragraph{Representing the derivative operator}
As an example, consider discretizing the derivative operator $D = \partial/\partial x$ into $\mathbf{\hat D}$. Discretizing it over $\mathcal{F}$, which corresponds to a Fourier spectral method, boils down to applying the derivative rule of the Fourier transform over the parameters:
\begin{equation}
    \gamma = D_p(\mathcal{F})(\theta) = \text{iDFT}(-\rmi k_x\cdot\text{DFT}(\theta)),
\end{equation}
where $\rmi=\sqrt{1}$. The discretization family is still $\mathcal{F}$, that is $D_f(\mathcal{F})=\mathcal{F}$, and the discretized operator is then
\begin{equation}
    \mathbf{\hat D}(\mathcal{F}, \theta)  = (\mathcal{F},  D_p(\mathcal{F})(\theta)).
\end{equation}
Applying the same operator on the family $\mathcal{P}_N$ gives instead
\begin{equation}
    D \mathcal{P}_N(\theta,x) = \frac{\partial }{\partial x}\sum_{i=0}^N \theta_i x^i = \sum_{i=1}^{N-1}i\theta_ix^{i-1}.
\end{equation}
The parameter function  $D_p(\mathcal{P}_N)$ is
\begin{equation}
    D_p(\mathcal{P}_N)(\theta) = \Big(\theta_1, 2\theta_2, 3\theta_3,\dots,(N-1)\theta_{N-1}\Big),
\end{equation}
while $D_f\left(\mathcal{P}_N\right)$ maps to the space of polynomials with one degree less
\begin{equation}
    D_f\left(\mathcal{P}_N\right)(\theta, x) = \sum_{i=0}^{N-1} \theta_i  x^i.
\end{equation}
Note that here not only the parameters are changed, but also the discretization family. Doing the same procedure on a neural network to obtain a PINN leaves the parameters unchanged, while the output discretization family is obtained by applying AD on the interpolation function of the input. For a spline representation, the discrete parameters are processed by filtering with an FIR filter, while the discretization family is given by splines of a lower degree on a staggered collocation grid \cite{unser1999splines}.

In general, one may want to control both the choice of discretization family and the corresponding representation of various operators. For example, the product between two functions in $\mathcal{F}$ should be described over a denser collocation grid, to account for the larger spectral support of the output \cite{luise2009teoria}; however, when working with low-frequency signals an acceptable approximation is given by maintaining the original discretization, and simply multiplying element-wise the parameters on the two original spatial grids.

\paragraph{Binary operators}
Binary operators $\mathbf{B}(f,g) = h$ that take two fields can be defined analogously to the standard operators above, as 
\begin{equation}
    \mathbf{\hat B}(\mathcal{D},\theta_f,\theta_g) = \left(B_f(\mathcal{D}), B_p(\mathcal{D})(\theta_f,\theta_g)\right),
\end{equation}
assuming that $f$ and $g$ share the same discretization family $\mathcal{D}$.

\paragraph{Composing operators}
A nice characteristic of defining operators in this way can be seen when two operators are composed together. Given two operators $\mathbf{L}$ and $\mathbf{M}$, then the discretization of the composition operator $\mathbf{H}(f) = \mathbf{M} \circ \mathbf{L} (f) = \mathbf{M}(\mathbf{L}(f))$ is found as
\begin{align}
    \mathbf{\hat{L}}(\mathcal{D},\theta) & = (\mathcal{E}, \gamma) & = &(L_f(\mathcal{D}), L_p(\mathcal{D})(\theta)) \\
    \mathbf{\hat M}\circ \mathbf{\hat L}(\mathcal{D},\theta) & = (M_f(\mathcal{E}), M_p(\mathcal{E})(\gamma)) 
        & = &\Big(M_f(L_f(\mathcal{D})), M_p(L_f(\mathcal{D}))(L_p(\mathcal{D})(\theta))\Big)
\end{align}
This shows that the final discretization family $M_f(L_f(\mathcal{D}))$ and the function which transforms the parameters are simply given by function compositions, and can be found ahead of computation once $\mathcal{D}$ is known. Importantly, if $M_p(L_f(\mathcal{D}))$ and $L_p(\mathcal{D})$ are written in a differentiable language, then $H_p(\mathcal{D})$ can also be differentiated. We choose to implement them in JAX \cite{jax2018github}.

\paragraph{Operator parameters}
Operators often have parameters associated with them, similarly to neural network layers. As an example, discretizing the derivative operator with finite differences corresponds to the following mapping (ignoring boundary conditions):
\begin{equation}
    \mathbf{\hat D}(\mathcal{Q}, \theta) = \left(\mathcal{Q}, w*\theta\right),    
\end{equation}
where $\mathcal{Q}$ is the Finite Differences (FD) discretization family, $*$ denotes convolution and $w \in R^K$ is an appropriate kernel with $K$ elements. We may want to explicitly learn the discretized PDE from data \cite{long2018pde}, meaning that we must be able to find the gradient with respect to $w$. For a general operator $\mathbf{L}$, we account for this possibility by allowing $L_p$ to accept an extra appropriately initialized parameter, such that
\begin{equation}
    \mathbf{L}(\mathcal{D},\theta) = (L_f(\mathcal{D}), L_p(\mathcal{D})(\theta, w)).
\end{equation}
Finally, we may want to share some parameters between different operators that are computing similar quantities. Those extra parameters complicate the composition rules, but this can be accounted for by appropriately tracing and initializing the various parameters when building the computational graph of $L_p$.

\subsection{Software abstraction}
In the proposed software, fields are represented as couples $(\mathcal{D}, \theta)$ made of a discretization family and set of parameters. Each discretization must define its own interpolation function. They are subclasses of a common base discretization, with operators defined as class methods using function composition and AD: this ensures that most operators work on arbitrary user-defined discretizations, albeit not necessarily in the most efficient way.

Specific implementations of new or existing operators can be created by overriding the corresponding method, by providing the rules for evaluating $L_f$ and $L_p$, as well as the initialization of the operator parameters $w$.
The software discretizes the operator by building and tracing a lower level computational graph, translating every single operation in the corresponding JAX function for the given discretization. The output is a set of discrete parameters, collected in a dictionary, and the sampling operator, as a pure JAX function. The latter can be differentiated, jit-compiled, parallelized and manipulated using JAX program transformations. The resulting discretization also natively runs on GPUs and TPUs.

In its essence, this library works in a similar way as most machine learning libraries currently available: using self-contained composable operations as computational layers, where each module depends on a set of parameters and an input. The key difference is that each operation represents an abstract mathematical mapping, which is replaced by an appropriate algorithm depending on the input discretization. This decouples the mathematical definition of the equation from the actual implementation. For example, the operation {\tt gradient} can be implemented using AD on the interpolation function, as is done in PINNs, or by a convolution operator as in FD, or by an elementwise product in the Fourier domain.
Another source of difference to standard ML libraries, is that different operators may share common parameters, if required by the user.
Lastly, it is easy to customize or write new discretizations, as well as building hierarchical families of discretizations, using inheritance and method overloading.

\section{Numerical examples}

\subsection{Differentiability}
To demonstrate differentiability and automatic discretization, we apply the framework to the solution of a time-harmonic acoustic optimization problem, where the physics is described by the Helmholtz equation with frequency $\omega=1$. The goal is to maximize the complex wavefield amplitude at a target location $x_T$. The source is given by an omnidirectional point transducer. The speed of sound map $c(\rho)$ is parametrized in the central region by an array of real values $\rho$, constrained between 1 and 2 using a sigmoid function. The array $\rho$ is initialized randomly and $c=1$ outside the central region.

The optimization problem requires the following function to be minimized:
\begin{equation}
    L = - \|u(x_T)\| + \lambda \text{TV}(c(\rho)), \quad s.t. \left(\nabla^2 + \frac{\omega^2}{c^2} \right)u = -S_M,
\end{equation}
where $u$ is the solution of the Helmholtz equation, $S_M$ is the complex source field, $\lambda$ is an arbitrary positive constant set to $10^{-4}$ and $\text{TV}(c(\rho))$ is a Total-Variation regularization term, defined as
\begin{equation}
    \text{TV}(c) = \frac{1}{\|\Pi\|}\int_\Pi \|\nabla c\| \,dx.
\end{equation}

The derivative operators need to be modified at the boundary, to enforce radiating boundary conditions, using a Perfectly Matched Layer as:
\begin{equation}
    \nabla^2 = \sum_{j=(x,y)}\hat\partial_{j}^2, \qquad \hat \partial_{j} = \frac{\partial_{j}}{\gamma_j(x)},
\end{equation}
where $\gamma_j$ are appropriate complex fields \cite{bermudez_optimal_2007}.

\paragraph{Discretization}
This equation can then be expressed using a language very close to the mathematical formulation in the proposed library. We define two operators: the Helmholtz operator and the integrand of the TV functional: 
\begin{minted}[fontsize=\footnotesize]{python}
from jaxdf import operator
from jaxdf import operators as jops

@operator()
def helmholtz(u, c, x):
    pml = gamma(x)
    mod_grad_u = jops.gradient(u)*pml
    mod_diag_jacobian = jops.diag_jacobian(mod_grad_u)*pml
    laplacian = jops.sum_over_dims(mod_diag_jacobian)
    return laplacian + ((1./c)**2)*u

@operator()
def integrand_TV(u):
    nabla_u = jops.gradient(u)    
    return jops.sum_over_dims(jops.elementwise(jnp.abs)(nabla_u))
\end{minted}

This operator can be discretized arbitrarily, by calling it with a field defined over a discretization family. For example, if we aim to use a feed-forward neural network to represent $u$, we can do so by defining a new {\tt{Arbitrary}} discretization where the interpolation function is defined using the neural network forward function.
If instead one wants to solve the equation using Fourier spectral methods, as we will do in this example, it suffices to instantiate a {\tt FourierSeries} discretization and call the operators with the corresponding fields. The full code for this experiment is listed in Appendix \ref{sec:code_helmholtz_experiment}.

\paragraph{Optimization}

\begin{figure}
    \makebox[\textwidth][c]{\hspace*{-2.2cm}
        \scalebox{0.64}{\input{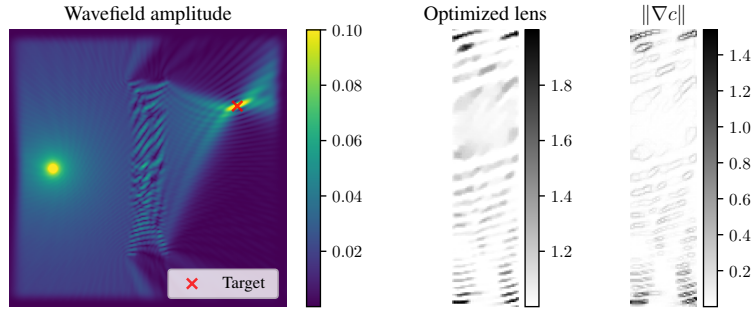}}
    }
    \caption{Wavefield, lens and total variation integrand after optimization. \label{fig:optimized_lens}}
\end{figure}

With the operator discretized, we are equipped with two differentiable pure functions (see Appendix \ref{sec:code_helmholtz_experiment}). Such functions accept as input the discrete parameters of $u$ and $c$, as well as the parameters required by the intermediate operators. 
Because a Fourier discretization preserves the linearity of the Helmholtz operator, the acoustic field can be found using the Generalized minimal residual method (GMRES), while gradients are computed with constant memory size via the Implicit Function Theorem \cite{blondel2021efficient}. The loss function is minimized using 100 steps of the Adam optimizer, with the learning rate set to 0.1 and $\beta$s set to (0.9, 0.9).

The result of the optimization is shown in Fig. \ref{fig:optimized_lens}, showing that the speed of sound of the lens has been successfully optimized to provide focus at the location of interest.

\subsection{Swapping discretization}
A key feature of the proposed framework is the ability to quickly change discretization for a given operator. To demonstrate this, we'll focus on solving the initial value problem for the heat equation

\begin{equation}
    u = \int_{0}^t\nabla^2 u(\tau) \, d\tau,\qquad \text{s.t. } u(0) = u_0
\end{equation}

Note that we don't enforce specific boundary conditions here. In the current version of the framework, boundary conditions are not yet implemented and are implicitly defined by the discretization (e.g. periodic for Fourier Series, zero-padding for Finite Differences etc.). This is not suitable for proper integration with arbitrary boundary conditions, but works for equations where it is feasible to construct an absorbing layer at the boundary: that is the majority of situations, for example, when investigating wave phenomena. While the user is always free to use a custom discretization that satisfies the boundary conditions by construction, more suitable handling of generic boundary conditions would be a great addition to the package and will be part of future work.

Boundary conditions aside, the code for defining and discretizing the operator using Finite Differences is the following:

\begin{minted}[fontsize=\footnotesize]{python}
@operator()
def heat_rhs(u):
    return jops.laplacian(u)

# Discretize using Finite Differences\
discr = RealFiniteDifferences(domain, accuracy=2)
u_params, u = discr.empty_field(name='u')

# Construct numerical function and collect parameters
RHS_FD = heat_rhs(u=u)
num_op = RHS_FD.get_field_on_grid(0)
global_parameters = RHS_FD.get_global_params()
\end{minted}

Suppose we now want to switch to a Fourier Series discretization, rather than Finite Differences. To reuse the previously defined operator, we need to perform three steps: define a new field with the new discretization, call the operator and extract the parameters and numerical function

\begin{minted}[fontsize=\footnotesize]{python}
# New Field
discr = RealFourierSeries(domain)
u_params, u = discr.empty_field(name='u')
u0 = u_params.at[...,0].set(img)

# Compile operator, generates the new discretization
RHS_Fourier = heat_rhs(u=u)

# Get parameters and pure function
gp = RHS_Fourier.get_global_params()
num_op = RHS_Fourier.get_field_on_grid(0)
\end{minted}

\begin{figure}[tp]
    \begin{center}
        \includegraphics[width=\textwidth]{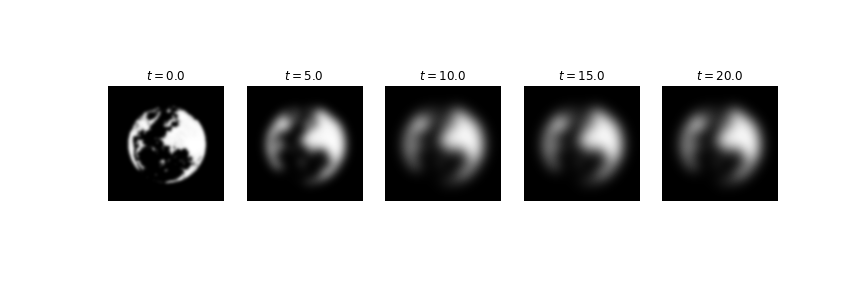}
    \end{center}
    \caption{Integration of the heat equation for an arbitrary initial field, using a Fourier spectral discretization for the Laplacian. \label{fig:heat_eq}}
\end{figure}

The result of integrating the heat equation for a given initial field is shown in Fig. \ref{fig:heat_eq}, where the Laplacian operator has been discretized using Fourier spectral methods.

\section{Conclusions}
In this paper, we have proposed a design framework for using and implementing custom discretizations, employed to compute the effect of operators using differentiable functions. New discretizations can be easily defined, by providing the corresponding interpolation function. The behaviour of each operator, for a given discretization, can be specified by the user or automatically evaluated using AD and function composition over the interpolation function. Future work will implement boundary conditions, which can still be seen as the mapping between different families of functions with the caveat of \textit{reducing} the number of free parameters, and the conversion from one discretization to another within the same operator, for example by defining projection operators.
{
\small
\printbibliography
}


\newpage
\appendix

\section{Full code for the Helmholtz example}
\label{sec:code_helmholtz_experiment}

\begin{minted}[fontsize=\footnotesize]{python}
from jaxdf import operators as jops
from jaxdf.core import operator, Field
from jaxdf.discretization import FourierSeries, Coordinate
from jaxdf.geometry import Domain
from jaxdf.utils import join_dicts
from jax import numpy as jnp
from jax.scipy.sparse.linalg import gmres
from jax.experimental import optimizers
import jax

# Settings
domain = Domain((256, 256), (1., 1.))
seed = jax.random.PRNGKey(42)

# Speed of sound parametrization
lens_params = jax.random.uniform(seed, (168,40)) - 4
def get_sos(p):
    lens = jnp.zeros(domain.N).at[44:212,108:148].set(jax.nn.sigmoid(p)) + 1
    return jnp.expand_dims(lens, -1)

# Defining operators
@jops.elementwise
def pml_absorption(x):
    abs_x = jnp.abs(x)
    return jnp.where(abs_x > 110, (jnp.abs(abs_x-110)/(128. - 110)), 0.)**2

gamma = lambda x: 1./(1 + 1j*pml_absorption(x))

@operator()
def helmholtz(u, c, x):
    pml = gamma(x)
    mod_grad_u = jops.gradient(u)*pml
    mod_diag_jacobian = jops.diag_jacobian(mod_grad_u)*pml
    laplacian = jops.sum_over_dims(mod_diag_jacobian)
    return laplacian + ((1./c)**2)*u

@operator()
def integrand_TV(u):
    nabla_u = jops.gradient(u)    
    return jops.sum_over_dims(jops.elementwise(jnp.abs)(nabla_u))

# Defining discretizations
fourier_discr = FourierSeries(domain)
u_fourier_params, u = fourier_discr.empty_field(name='u')
src_fourier_params, src = fourier_discr.empty_field(name='src')
src_fourier_params = u_fourier_params.at[128, 40].set(1. + 0j)  # Monopole source
_, c = fourier_discr.empty_field(name='c')
_, x = fourier_discr.empty_field(name='x')
x_params = Coordinate(domain).get_field_on_grid()({}) # Coordinate field

# Discretizing operators: getting pure functions and parameters
H = helmholtz(u=u, c=c, x=x)
TV = integrand_TV(u=u)
global_params = join_dicts(H.get_global_params(), TV.get_global_params())
H_on_grid = H.get_field_on_grid(0)
tv_on_grid = lambda x: TV.get_field_on_grid(0)(global_params, {"u": x})

# Helmholtz solver function
def solve_helmholtz(speed_of_sound):
    params = {"c":speed_of_sound, "x":x_params}
    def helm_func(u):
        params["u"] = u
        return H_on_grid(global_params, params)
    sol, _ = gmres(helm_func, src_fourier_params, maxiter=1000, tol=1e-3, restart=50)
    return sol

# Loss function
def loss(p):
    sos = get_sos(p)
    tv_term = jnp.mean(tv_on_grid(sos))
    field = solve_helmholtz(sos)
    return -jnp.sum(jnp.abs(field[70,210])) + 1e-4*tv_term

# Optimization loop
init_fun, update_fun, get_params = optimizers.adam(.1, b1=0.9, b2=0.9)
opt_state = init_fun(lens_params)

@jax.jit
def update(opt_state, k):
    lossval, gradient = jax.value_and_grad(loss)(get_params(opt_state))
    return lossval, update_fun(k, gradient, opt_state)

for k in range(100):
    lossval, opt_state = update(opt_state, k)
    print(f"Step {k}, Loss {lossval}")
\end{minted}

\end{document}